\documentclass[lettersize,journal]{IEEEtran}
\usepackage{algorithmic}
\usepackage{algorithm}
\usepackage{array}
\usepackage[caption=false,font=normalsize,labelfont=sf,textfont=sf]{subfig}
\usepackage{textcomp}
\usepackage{stfloats}
\usepackage{url}
\usepackage{verbatim}
\usepackage{graphicx}
\usepackage{cite}
\usepackage{amsmath,amssymb,amsfonts}
\usepackage{booktabs}
\usepackage{multirow}
\usepackage{hyperref}
\usepackage{mathtools}
\usepackage{leftindex}
\usepackage{bbding}
\usepackage[table,xcdraw]{xcolor}
\begin{document}

\title{EndoGMDE: Generalizable Monocular Depth Estimation with Mixture of Low-Rank Experts for Diverse Endoscopic Scenes}
\author{Liangjing Shao, \IEEEmembership{Graduate Student Member, IEEE}, Chenkang Du, Benshuang Chen, Xueli Liu, Xinrong Chen
\thanks{This work was supported by National Natural Science Foundation of China (No.82472116), Natural Science Foundation of Shanghai (No.24ZR1404100) and Key Research and Development Plan of Ningxia Hui Autonomous Region (No.2023BEG02035).}
\thanks{Liangjing Shao is currently with Department of Electronic Engineering, The Chinese University of Hong Kong. This work was done when he was a master student with College of Biomedical Engineering, Fudan University.}
\thanks{Chenkang Du, Benshuang Chen and Xinrong Chen are with College of Biomedical Engineering, Fudan University, and also with Shanghai Key laboratory of Medical Image Computing and Computer Assisted Intervention.}
\thanks{Xueli Liu is with ENT Institute and Department of Otolaryngology, Eye \& ENT Hospital of Fudan University.}
\thanks{Corresponding Authors: Xueli Liu(liuxueli@fudan.edu.cn), Xinrong Chen(chenxinrong@fudan.edu.cn)}}

\maketitle

\begin{abstract}
Self-supervised monocular depth estimation is a significant task for low-cost and efficient 3D scene perception and measurement in endoscopy. However, the variety of illumination conditions and scene features is still the primary challenges for depth estimation in endoscopic scenes. In this work, a novel self-supervised framework is proposed for monocular depth estimation in diverse endoscopy. Firstly, considering the diverse features in endoscopic scenes with different tissues, a novel block-wise mixture of dynamic low-rank experts is proposed to efficiently finetune the foundation model for endoscopic depth estimation. In the proposed module, based on the input feature, different experts with a small amount of trainable parameters are adaptively selected for weighted inference, from low-rank experts which are allocated based on the generalization of each block. Moreover, a novel self-supervised training framework is proposed to jointly cope with brightness inconsistency and reflectance interference. The proposed method outperforms state-of-the-art works on SCARED dataset and SimCol dataset. Furthermore, the proposed network also achieves the best generalization based on zero-shot depth estimation on C3VD, Hamlyn and SERV-CT dataset. The outstanding performance of our model is further demonstrated with 3D reconstruction and ego-motion estimation. The proposed method could contribute to accurate endoscopy for minimally invasive measurement and surgery. The evaluation codes have been released on \href{https://github.com/BaymaxShao/Endo-GMDE}{https://github.com/BaymaxShao/Endo-GMDE}, while the demo videos can be found on \href{https://endo-gmde.netlify.app/}{https://endo-gmde.netlify.app/}.
\end{abstract}


\section{Introduction}
Endoscopy, as one of primary minimally invasive surgical procedures, is widely used in the medical treatment due to its minor additional trauma and short recovery period. However, due to the limited field of view and 2D vision of endoscopes, 3D scene measurement has become a significant task for the precise perception of anatomical structures, such as lesion size and vascular boundaries.\cite{bg,mis} Depth estimation, as one of the most common 3D vision techniques which has been applied in underwater scenes\cite{uw1,uw2}, outdoor driving scenes\cite{out1,out2} and indoor scenes\cite{in}, has also been a primary 3D scene perception method in minimally invasive surgery\cite{mis}.

\begin{figure}
    \centering
    \includegraphics[width=\linewidth]{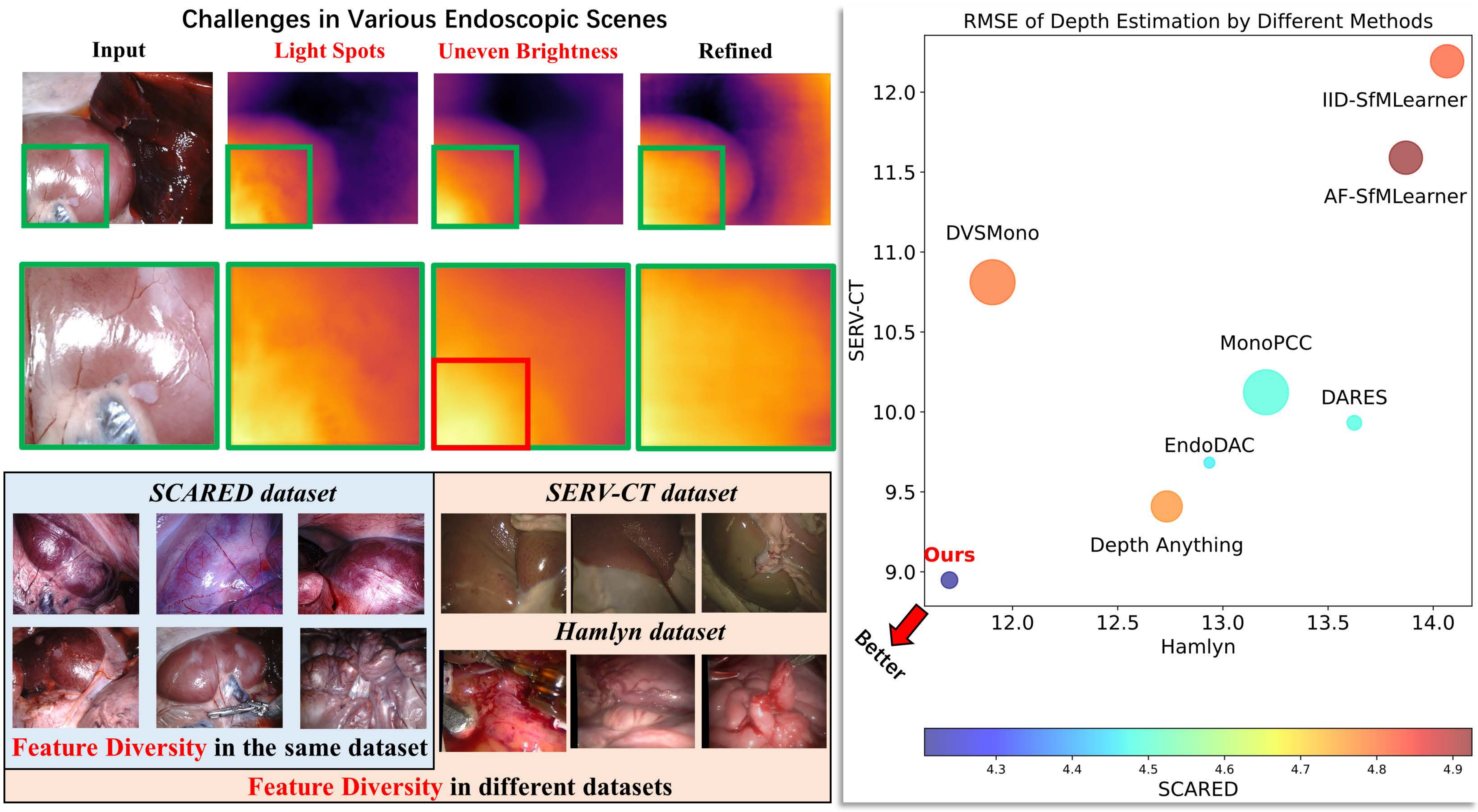}
    \caption{Challenges (left) and performance comparison (right) for depth estimation in diverse endoscopic scenes. The size of the circle denotes the number of trainable parameters in the corresponding depth estimation network.}
    \label{teaser}
\end{figure}

Due to the high cost of obtaining depth maps \cite{tmi} and limited inference efficiency of stereo matching \cite{sm} in endoscopy, self-supervised monocular depth estimation is the main-stream method for endoscopic scenes. Although numerous self-supervised methods, such as MonoViT\cite{mv} and LiteMono\cite{lm}, have been proposed for monocular depth estimation for natural scenes based on deep learning, there are two critical challenges for the application of such methods in endoscopy. Firstly, illumination conditions in endoscopic scenes often causes brightness and reflectance inconsistency. Moreover, the performance and generalization of the depth estimation model will be highly limited by the variety of scene features due to different surgical tasks and tissues, which is shown in Fig. \ref{teaser}.

To deal with the illumination inconsistency, Shao et al.\cite{afsfm} propose to utilize constraint based on the appearance flow to alleviate brightness inconsistency. Meanwhile, intrinsic image decomposition is also a useful technique to filter the interference of the light, which is also applied in endoscopy\cite{iidsfm}. However, as Fig. \ref{teaser} shows, the depth at some positions with high brightness on the surface will be estimated as a higher value than the ground truth, while some light spots from reflectance on the surface will highly influence the depth value at the corresponding position. With the development of foundation models for depth estimation, a series of methods are proposed to efficiently finetune the foundation model for adaptation to the endoscopic scenes, including Surgical-DINO\cite{sdino}, EndoDAC\cite{endodac} and DARES\cite{dares}. However, most of the existing finetuned foundation models for endoscopic depth estimation are based on Low-Rank Adaptation (LoRA)\cite{lora}, the performance of which may be constrained by the feature diversity in diverse endoscopic scenes.

To this end, a novel self-supervised framework, EndoGMDE, is proposed to boost the performance and generalization of monocular depth estimation in diverse endoscopic scenes. First of all, in the proposed framework, an end-to-end training pipeline based on the integration of a novel intrinsic image registration and brightness constraint \cite{afsfm} is utilized to jointly deal with brightness inconsistency and light interference. Specifically, the invariance of the reflectance image to illumination changes and the strong correlations in the gradient domain\cite{grad} are considered in the proposed intrinsic image decomposition. Furthermore, a novel parameter-efficient finetuning method, Block-wise Mixture of Low-Rank Experts (BW-MoLE), is proposed to finetune the foundation model for adaptive depth estimation in endoscopy based on diverse scene features, which is built upon LoRA and Mixture of Experts (MoE). Note that the low-rank experts are allocated based on the training quality of different Transformer blocks in the encoder of foundation model. The main contribution of this work can be summarized as the following:
\begin{enumerate}
    \item A novel parameter-efficient finetuning method with block-wise mixture of dynamic low-rank experts is proposed to adaptively utilize the low-rank experts for depth estimation based on diverse features, which can boost the generalization of the model.
    \item A novel self-supervised framework for depth estimation is proposed to jointly deal with the inconsistency of brightness and reflectance in endoscopic scenes, which can benefit the depth estimation from observations with diverse illumination conditions.
    \item The proposed method is evaluated by self-supervised depth estimation on SCARED\cite{scared} and SimCol\cite{simcol} dataset, as well as zero-shot depth estimation on C3VD\cite{c3vd}, Hamlyn\cite{edm} and SERV-CT\cite{servct} dataset. Our method achieves state-of-the-art performance compared with existing works. Moreover, the proposed method is applied to sim-to-real test\cite{em} and 3D reconstruction. 
\end{enumerate}

\section{Related Work}
In this section, the related works for self-supervised monocular depth estimation in endoscopy are reviewed, as well as the application of foundation models in endoscopic depth estimation.

\subsection{Self-supervised Depth Estimation for Endoscopy}
Based on the method proposed by Zhou et al.\cite{zhou}, Turan et al. \cite{turan} proposed the first self-supervised method for depth estimation in endoscopic scenes. In the next few years, some works for self-supervised depth estimation in endoscopy are also proposed with the employment of attention mechanism. Ozyoruk et al. \cite{endoslam} introduced spatial attention blocks into the depth estimation network and ego-motion estimation network, while Liu et al. \cite{cmpb2023} further integrated the spatial attention and channel attention in the proposed framework. In the most recent work, Yang et al. \cite{tmi} proposed a lightweight self-supervised depth estimation method for endoscopy with the combination of convolution and self-attention. Different from direct estimation from the observation, Liu et al. \cite{s2d1}, \cite{s2d2} proposed a novel strategy to perform depth estimation based on a depth completion network, which reconstructs dense depth maps based on RGB data and corresponding sparse depth maps. Although a large amount of methods have been proposed for self-supervised depth estimation, the application of these methods in endoscopy is still limited by illumination issues in the endoscopic images.

In 2022, Shao et al. \cite{afsfm} proposed AF-SfMLearner to firstly tackle the brightness inconsistency using a generalized dynamic image constraint based on the appearance flow in endoscopic scenes. Based on this, Li et al. \cite{iidsfm} recently proposed a self-supervised method based on intrinsic image decomposition to deal with the challenge of reflectance on the surface of the tissue for endoscopic depth estimation. Differently, Zhou et al. \cite{dvs} proposed a method to select candidate frames with lowest temporal variance and further assign the source frame with the highest view consistency. The most recent work, MonoPCC \cite{pcc}, utilized a novel photometric-invariant cycle constraint to obtain cycle-warped image from the target image to solve the problem of brightness inconsistency. In this work, a novel self-supervised training strategy is proposed to jointly deal with the illumination inconsistency and the reflectance interference in diverse endoscopic scenes.

\subsection{Foundation Models for Endoscopic Depth Estimation}
With the development of the foundation models for depth estimation, such as DINO\cite{dino}, series of Depth Anything\cite{da},\cite{da2}, the application of such models in depth estimation for various scenes has gained increasing attention. For example, Yu et al.\cite{vfm} utilize DINOv2 to integrate semantic and geometric features for depth estimation in outdoor scenes. Some methods for depth estimation in endoscopy based on foundation models are also proposed in recent years. Firstly, Cui et al. \cite{sdino} proposed to finetune DINO with Low-Rank Adaptation(LoRA) for depth estimation in endoscopic scenes. Based on this, a self-supervised framework, EndoDAC \cite{endodac}, based on Depth Anything model \cite{da} with the novel parameter-efficient finetuning strategy named as dynamic-vector LoRA (DV-LoRA) and inserted convolution block is proposed to perform endoscopic depth estimation with generalizability. Meanwhile, Zeinoddin et al. \cite{dares} also proposed a different finetuning strategy named vector-LoRA which is utilized on the Depth Anything model. In this work, the foundation model for depth estimation in diverse endoscopic scenes is proposed with a novel parameter-efficient finetuning method based on the integration of MoE and LoRA, which can strengthen the adaptation of the model to different scene features.

\begin{figure*}
    \centering
    \includegraphics[width=\linewidth]{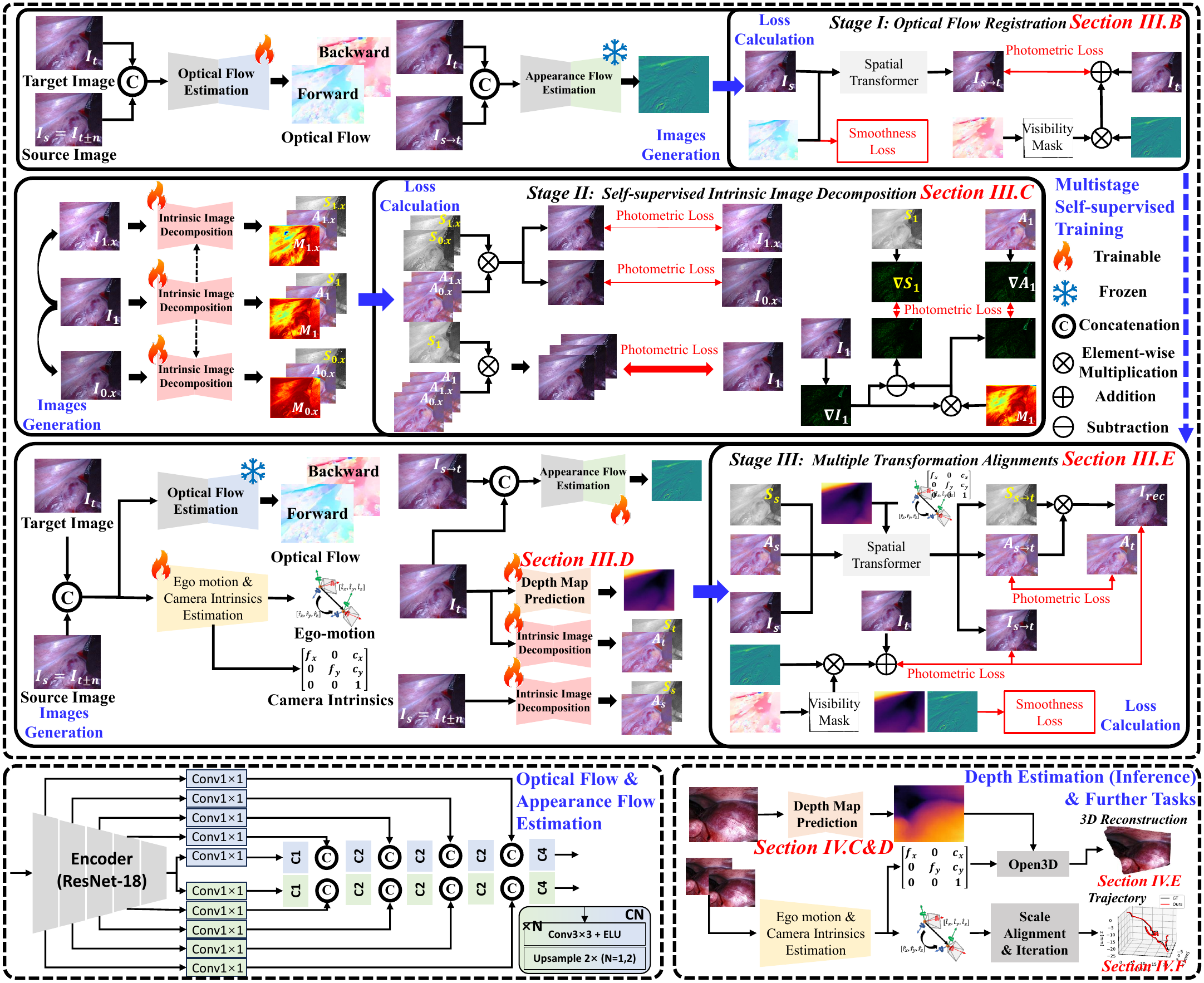}
    \caption{The overview of this work and the pipeline of the proposed self-supervised training.}
    \label{pipeline}
\end{figure*}

\section{Proposed Method}
\subsection{Overview}
The overview of the proposed method is shown as Fig. \ref{pipeline}. The proposed framework consists of five networks, which respectively perform optical flow prediction, appearance flow prediction, depth map prediction, intrinsic image decomposition, ego-motion estimation and camera intrinsic estimation. The framework of depth map prediction is a finetuned encoder-decoder network based on Depth Anything\cite{da}, which generates the corresponding depth map from single input image. Moreover, the ego-motion and camera intrinsics estimation network is mainly based on the module in \cite{endodac}.

The self-supervised learning of the framework is divided into three stages in each end-to-end training epoch to reserve the computational resources and pretend the information interference. The optical flow estimation network is firstly trained based on self-supervised pipeline introduced in Section \ref{s1}. Then, the intrinsic image decomposition network is trained based on the consistency between the recomposed images and the raw images at the second stage, detail of which is introduced in Section \ref{s2}. At the final stage, the depth map prediction network is efficiently finetuned by the proposed block-wise mixture of low-rank experts in Section \ref{peft}. Meanwhile, other networks, except the optical flow estimation network trained at the first stage, are fully trained at the final stage based on the alignments of rigid transformations using estimated ego-motion and depth map, stated in Section \ref{s3}. The photometric loss is utilized to supervise the consistency of images, defined as Eq. \ref{lp}.
\begin{equation}
    \mathcal{L}_p(\hat{I},I)= V(\beta\frac{1-SSIM(I,\hat{I})}{2}+(1-\beta)|I-\hat{I}|)
    \label{lp}
\end{equation}
where $V$ is the visibility mask generated from backward optical flow following \cite{afsfm}, $\beta=0.85$.

\subsection{Training Stage I: Optical Flow Registration} 
\label{s1}
Following \cite{afsfm}, the optical flow prediction network in Fig. \ref{pipeline} is optimized based on the loss function Eq. \ref{l1}, where $O_t$ is the forward optical flow of the source image $I_t$. $I^{opt}_{s\rightarrow t}$ is generated from the source image $I_t$ based on the forward optical flow and Spatial Transformer.
\begin{equation}
    \mathcal{L}_{1}=\mathcal{L}_p(I'_t,I^{opt}_{s\rightarrow t})+0.001|\nabla O_t|*e^{-\nabla |I_t|}
    \label{l1}
\end{equation}

\subsection{Training Stage II: Self-supervised Intrinsic Image Decomposition} 
\label{s2}

\begin{figure}
    \centering
    \includegraphics[width=\linewidth]{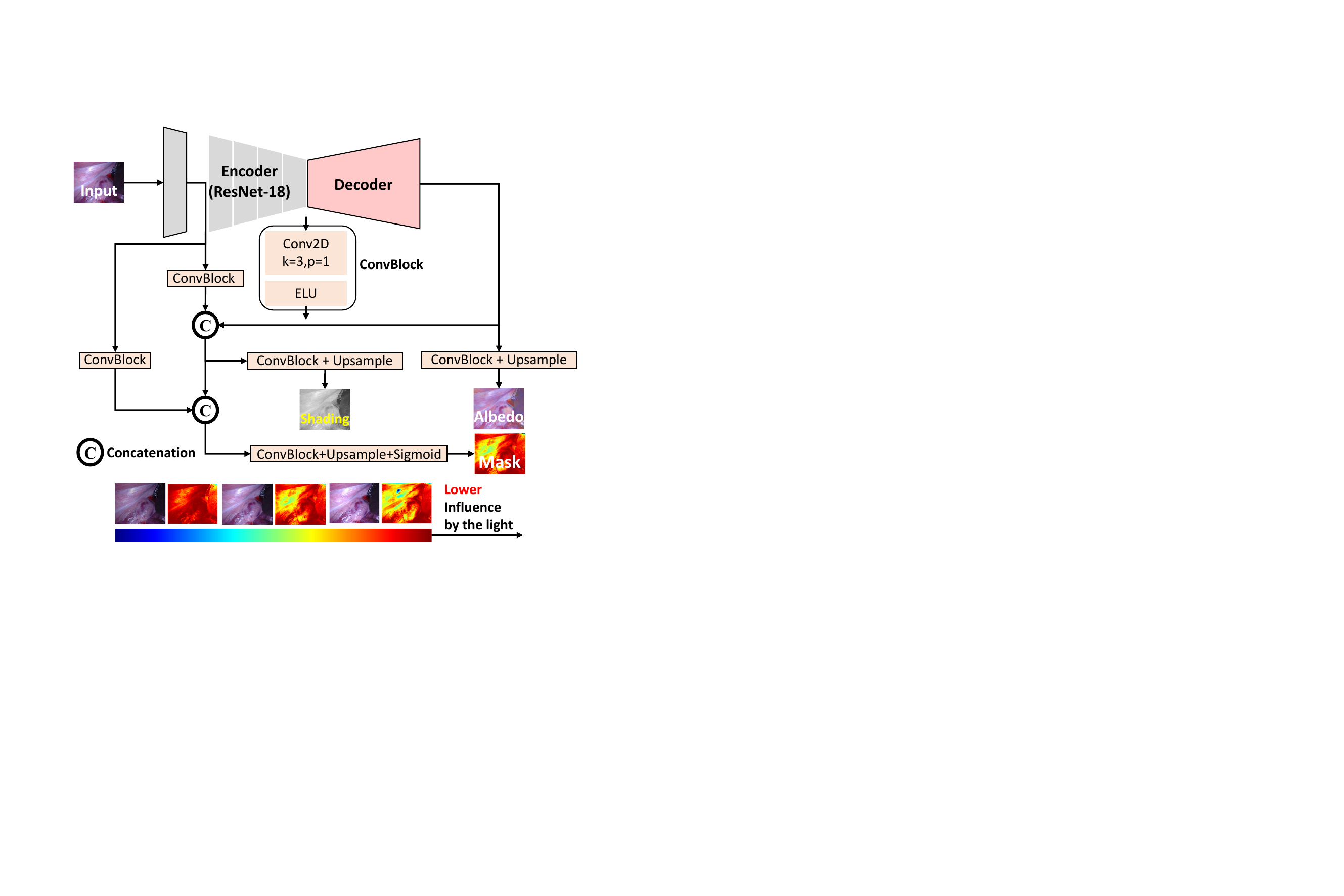}
    \caption{The framework for intrinsic image decomposition}
    \label{iid}
\end{figure}

The framework for intrinsic image decomposition shown in Fig. \ref{iid} is designed based on the encoder-decoder structure for optical flow prediction and appearance flow prediction networks shown in Fig. \ref{pipeline}. Given the input image $I$, the albedo image $A$, the shading image $S$ and the assignment mask $M$ are predicted by the intrinsic image decomposition network. The albedo image represents the inherent color and texture of a surface independent of illumination conditions, while the shading image represents the light conditions. The assignment mask represents the probability of each pixel derivation from albedo, which can denote the influence of the illumination. The relationship of the raw image, the albedo image and the shading image can be defined as $I = A * S$.

Firstly, the raw image will be randomly augmented with $n\in(0,2)$ times higher brightness as the image $I_{n\times}$. As the albedo image won't be influenced by illumination changes, the raw image could be recomposed by the albedo image $A_{1\times}$, $A_{n\times}$ from the raw image $I_{1\times}$ and the randomly augmented image $I_{n\times}$ and the shading image $S_{1\times}$ from the raw image, respectively. Therefore, the supervision of the recomposition is formulated as Eq. \ref{self}.

\begin{equation}
    \mathcal{L}_{rec}=\mathcal{L}_p(A_{1\times}*S_{1\times}, I_{1\times})+\mathcal{L}_p(A_{n\times}*S_{1\times}, I_{1\times})
    \label{self}
\end{equation}

Referring to \cite{grad}, there are strong correlations between the raw image and the intrinsic images at the gradient domain. Based on Retinex algorithm\cite{ret}, the decomposed image can be estimated from the raw image and the assignment mask. To this end, the photometric loss between the gradient of the estimated decomposed image and the gradient of the predicted image will be utilized to supervise the optimization of the intrinsic image decomposition, formulated as Eq. \ref{ret}.

\begin{equation}
    \mathcal{L}_{ret} = \mathcal{L}_p(\nabla A_{1\times},\nabla I_{1\times}*M_{1\times})+\mathcal{L}_p(\nabla S_{1\times},\nabla I_{1\times}*(1-M_{1\times}))
\label{ret}
\end{equation}

At the second stage, the intrinsic image decomposition network is solely optimized by the loss function $\mathcal{L}_2$ shown as Eq. \ref{l2} based on above losses.

\begin{equation}
    \mathcal{L}_{2}= \mathcal{L}_{rec}+0.1\mathcal{L}_{ret}
    \label{l2}
\end{equation}

\subsection{Parameter-efficient Finetuning for Depth Map Prediction}
\label{peft}

\begin{figure}
    \centering
    \includegraphics[width=\linewidth]{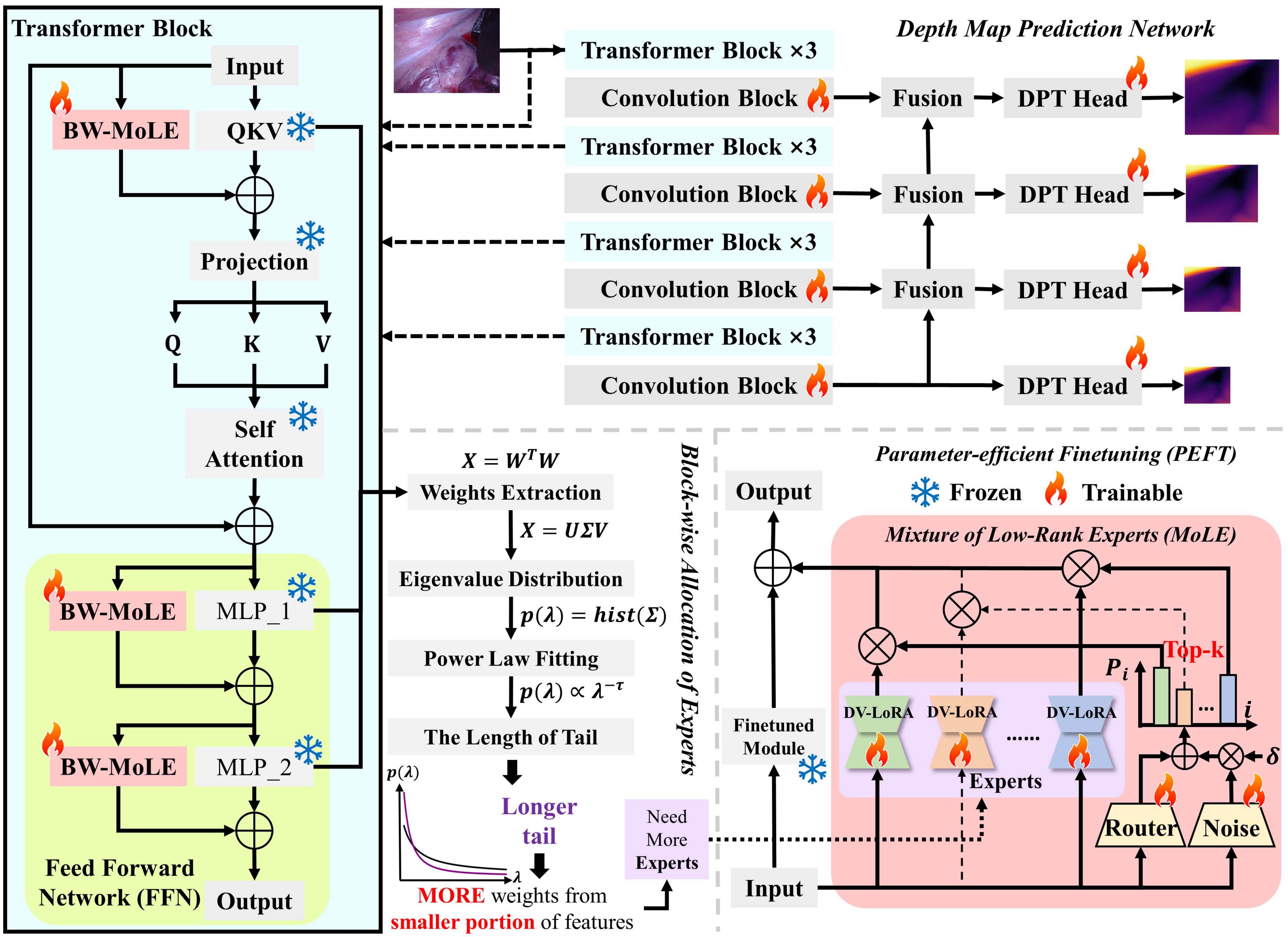}
    \caption{The proposed parameter-efficient finetuning of the depth map prediction network.}
    \label{fit}
\end{figure}

The depth map prediction network in the proposed framework is efficiently finetuned by block-wise mixture of dynamic low-rank experts (BW-MoLE), which is shown in Fig. \ref{fit}. Specifically, in each Transformer block of the network, the linear layer which generates query, value and key, as well as MLPs in the feed forward network (FFN), are finetuned by the proposed method.

\subsubsection{Mixture of Low-Rank Experts}
Based on DV-LoRA\cite{endodac}, multiple low-rank experts are implemented to update the output based on the frozen pretrained weights. Given the input feature, the route network $\mathcal{R}$ will predict the corresponding weights of the experts network by Eq. \ref{weight}, in which the noise generated by the network $\mathcal{N}$ is also introduced for robustness. The experts with top $k$ weights $\alpha_1,\alpha_2,...\alpha_k\in \Gamma_b=\{\alpha_1,\alpha_2,...,\alpha_{E_b}|\alpha_i>\alpha_{i+1}\}$ are selected to perform weighted finetuning. The finetuning process can be formulated as Eq. \ref{ft}.
\begin{equation}
    \Gamma=\mathcal{R}(X)+\delta\cdot\mathcal{N}(X)
    \label{weight}
\end{equation}
\begin{equation}
    Y'=Y+\sum_{i=1}^{k}\alpha_iV_iB_iU_iA_iX
    \label{ft}
\end{equation}
where $V_i$ and $U_i$ are trainable diagonal matrices of trainable weights $B$ and $A$, while $A,B$ are also trainable weights. Given the input feature map $X$, $Y$ is the output based on the frozen pretrained weights.

\subsubsection{Block-wise Allocation of Experts}
The weights of the network can be optimized by the smaller portion of specific features in the corresponding domain, which leads to limited generalization or transferring of the foundation model. Therefore, in this work, the number of experts is allocated based on the generalization of each block in the foundation model. To measure the generalizability, we analyze the distribution $\mathcal{D}(\lambda)$ of the eigenvalues $\lambda\in\Sigma^2$ based on the singular value decomposition $W=U\Sigma V$ of the extracted weights $W$ from linear layers, as Eq. \ref{dist} shows, where $\hat{\lambda}= argmax_{\lambda}(\log_{10}(\mathcal{D}(\lambda)))$. Based on the previous work\cite{alc}, the power law is utilized to fitting the probability distribution using the Hill estimator (Eq. \ref{hill}) and the Fix-finger \cite{ff} method (Eq. \ref{ffm}). 
\begin{equation}
    \rho(\lambda) = \log_{10}(\mathcal{D}(\lambda)) \varpropto \lambda^{-\tau}, 0.95\hat\lambda<\lambda<1.5\hat\lambda
    \label{dist}
\end{equation}
\begin{equation}
    \tau_j=1+\frac{N-j}{\sum_{i=1}^j\log_{10}\frac{\lambda_{N-i+1}}{\lambda_{j}}}, j=1,2,...,N-1
    \label{hill}
\end{equation}
\begin{equation}
    \tau=\tau_t, t= \underset{t}{argmin}(\max(|1-(\frac{\Lambda_t}{\lambda_t})^{-\tau_t+1}-\frac{C_t}{N-t}|))
    \label{ffm}
\end{equation}
where $\Lambda_t=[\lambda_t,\lambda_{t+1},...,\lambda_N]$, $N$ is the number of eigenvalues, $C_t=[0,1,2,..,N-t]$.

The power law distribution with larger $\tau$ means that the distribution of the eigenvalues from pretrained weights is with the longer tail, which denotes that more weights from the smaller portion of features. Therefore, if the distribution of the eigenvalues from the pretrained weights is with larger $\tau$, more experts are needed to finetune the model for generalization. To this end, Eq. \ref{allo} is utilized to allocate the number of experts for each block $E_b$ based on the distributions.
\begin{equation}
    E_b = \lfloor\frac{\tau_b}{\sum_{i=1}^B\tau_i}\times E_0\rfloor
    \label{allo}
\end{equation}
where $E_0$ is the total number of experts.

\subsection{Stage III: Multiple Transformation Alignments} 
\label{s3}
Given the predicted depth map $D_s$, ego-motion $P$ and camera intrinsics $K$, the rigidly transformed image $I_{s\rightarrow t}$ can be generated from $I_s$ based on Eq. \ref{rig} where $\leftindex^t{H}$ denotes the translation vector from the homogeneous matrix $H$, while $\leftindex^r{H}$ denotes the rotation matrix in $H$. $I(p)$ represents the homogeneous coordination of the pixel $p$ in the image $I$.
\begin{equation}
    [\leftindex^t{(KI_s(p))}, D_s(p)]=\leftindex^r{P}^{-1}([\leftindex^t{(KI_{s\rightarrow t}(p))}, D_t(p)]-\leftindex^t{P})
    \label{rig}
\end{equation}
Following Eq. \ref{rig}, the source images $I_t$ and their intrinsic images $A_t,S_t$ are all transformed into the pseudo images $I_{s\rightarrow t}$, $A_{s\rightarrow t}$ and $S_{s\rightarrow t}$. Based on the pseudo intrinsic images, the pseudo target images can be reconstructed and aligned with the target image $I_t$ by Eq. \ref{reco}. Meanwhile, the photometric loss between the albedo image of the target image and the pseudo albedo image shown as Eq. \ref{if} is utilized to supervise the rigid transformation, without influence from the illumination. Moreover, the smoothness of the depth map $D_t$ and the appearance flow $A_p$ is supervised by the loss function Eq. \ref{sm}.
\begin{equation}
    \mathcal{L}_{rec}=0.02\mathcal{L}_p(A_{s\rightarrow t}*S_{s\rightarrow t}, I_t)+0.01\mathcal{L}_p(I_{s\rightarrow t}, I_t)
    \label{reco}
\end{equation}
\begin{equation}
    \mathcal{L}_{if}=\mathcal{L}_p(A_{s\rightarrow t}, A_t)
    \label{if}
\end{equation}
\begin{equation}
    \mathcal{L}_{sm}=0.01|\nabla A_p| \cdot e^{\nabla|I_t - I^{opt}_{s\rightarrow t}|} + 0.001|\nabla D_t| \cdot e^{\nabla|I_t|}
    \label{sm}
\end{equation}
Based on the above losses, the final loss function for the optimization of the networks except optical flow prediction network is defined as Eq. \ref{l3}
\begin{equation}
    \mathcal{L}_3=\mathcal{L}_{sm}+\mathcal{L}_{rec}+0.02\mathcal{L}_{if}
    \label{l3}
\end{equation}

\section{Experiments and Results}

\begin{figure}
    \centering
    \includegraphics[width=\linewidth]{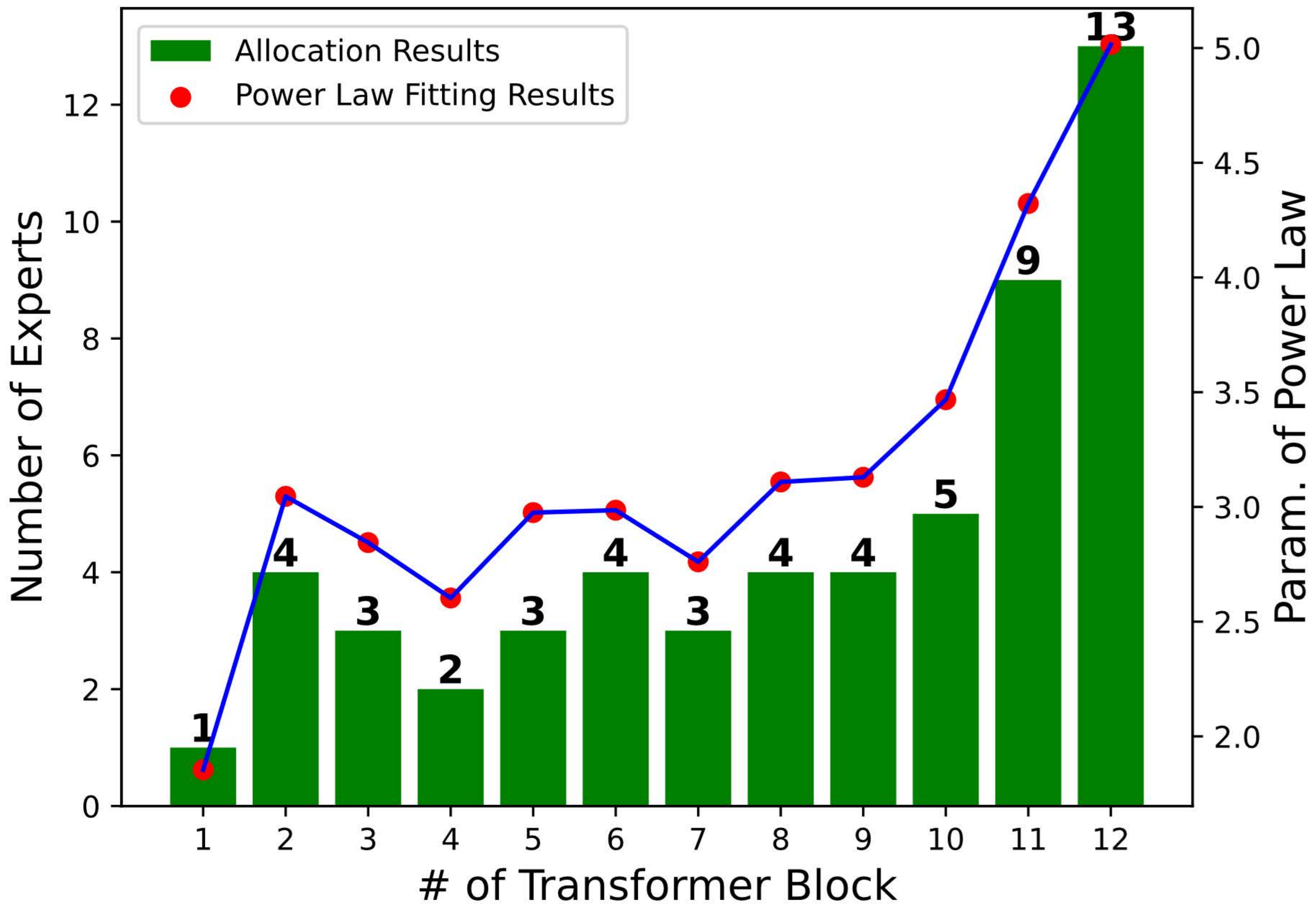}
    \caption{Allocation of experts. 'Param. of Power Law' denotes the $\tau$ in Eq. \ref{dist}}
    \label{allocation}
\end{figure}

\subsection{Implementation Details and Metrics}
The proposed method is implemented on Ubuntu 22.04 with one NVIDIA RTX 4090 GPU. The framework is trained for 20 epochs in total. For each training stage, a dependent Adam optimizer is set with initial learning rate of 1e-4, which will decayed to 1e-5 after 10 epochs. For datasets SCARED, Hamlyn and SERV-CT, all images are resized to 256$\times$320, while images from SimCol, C3VD and EndoMapper are resized to 256$\times$256. The rank of low-rank experts is set to 4, and the number of warm-up steps for DV-LoRA\cite{endodac} is set to 20000. The allocation of 55 low-rank experts to 12 Transformer blocks is shown as Fig. \ref{allocation}, and the top-k of each finetuning module is set to 2, which will be set to 1 if the number of experts equals to 1. The sampling interval $n$ between the target image $I_i$ and source images $I_{i\pm n}$ is set to 4. The evaluation metrics are defined as Eq. \ref{m1}, Eq. \ref{m2}, Eq. \ref{m3}, Eq. \ref{m4} and Eq. \ref{m5}, where $\mathbf{D}$ is the set of the predicted depth. $\hat{d}$ and $d$ denote the predicted depth and the corresponding ground truth, respectively.
\begin{equation}
    Rel_{Abs}=\frac{1}{|\mathbf{D}|}\sum_{\hat{d}\in\mathbf{D}}|d-\hat{d}|/d
    \label{m1}
\end{equation}
\begin{equation}
    Rel_{Sq}=\frac{1}{|\mathbf{D}|}\sum_{\hat{d}\in\mathbf{D}}\frac{|d-\hat{d}|^{2}}{d}
    \label{m2}
\end{equation}
\begin{equation}
    RMSE=\sqrt{\frac{1}{|\mathbf{D}|}\sum_{\hat{d}\in\mathbf{D}}|d-\hat{d}|^2}
    \label{m3}
\end{equation}
\begin{equation}
    RMSE_{Log}=\sqrt{\frac{1}{|\mathbf{D}|}\sum_{\hat{d}\in\mathbf{D}}|\log d-\log \hat{d}|^2}
    \label{m4}
\end{equation}
\begin{equation}
    \delta=\frac{1}{|\mathbf{D}|}\left|\left\{\hat{d}\in\mathbf{D}|max(\frac{d}{\hat{d}},\frac{\hat{d}}{d}<1.25)\right\}\right|
    \label{m5}
\end{equation}

\begin{table*}[]
\centering
\fontsize{9}{10}\selectfont
\renewcommand{\arraystretch}{1}
\caption{Comparison Results of Self-supervised Depth Estimation on SCARED. \textbf{The BEST results} are in BOLD. $\dagger$ denotes that the foundation model is finetuned based on \cite{endodac}. \textbf{TP} denotes the number of trainable parameters.}
\begin{tabular}{cccccccc}
\hline
\multicolumn{1}{c|}{\textbf{Methods}}         & \multicolumn{1}{c|}{\textbf{Year}}                & \textbf{$Rel_{Abs}\downarrow$} & \textbf{$Rel_{Sq}\downarrow$} & \textbf{$RMSE\downarrow$} & \textbf{$RMSE_{Log}\downarrow$} & \multicolumn{1}{c|}{\textbf{$\delta\uparrow$}}              & \textbf{TP/M} \\ \hline
\multicolumn{1}{c|}{AF-SfMLearner \cite{afsfm}}                                   & \multicolumn{1}{c|}{2022}                     & 0.059                          & 0.435                         & 4.925                     & 0.082                           & \multicolumn{1}{c|}{0.974}                                  & 14.8          \\
\multicolumn{1}{c|}{MonoViT \cite{mv}}                                       & \multicolumn{1}{c|}{2022}                       & 0.062                          & 0.470                         & 5.042                     & 0.082                           & \multicolumn{1}{c|}{0.976}                                  & 27.9          \\
\multicolumn{1}{c|}{LiteMono \cite{lm}}                                      & \multicolumn{1}{c|}{2023}                      & 0.057                          & 0.453                         & 4.967                     & 0.079                           & \multicolumn{1}{c|}{0.975}                                  & 3.1           \\
\multicolumn{1}{c|}{Surgical-DINO \cite{sdino}}                                    & \multicolumn{1}{c|}{2024}                    & 0.059                          & 0.427                         & 4.904                     & 0.081                           & \multicolumn{1}{c|}{0.974}                                  & 0.1           \\
\multicolumn{1}{c|}{Yang et al.\cite{tmi}}                                           & \multicolumn{1}{c|}{2024}                       & 0.062                          & 0.558                         & 5.585                     & 0.090                           & \multicolumn{1}{c|}{0.962}                                  & 2.0           \\
\multicolumn{1}{c|}{Depth   Anything$\dagger$\cite{da}}                            & \multicolumn{1}{c|}{2024}                      & 0.055                          & 0.410                         & 4.769                     & 0.078                           & \multicolumn{1}{c|}{0.973}                                  & 13.0          \\
\multicolumn{1}{c|}{Depth Anything   v2$\dagger$\cite{da2}}                          & \multicolumn{1}{c|}{2024}                   & 0.076                          & 0.683                         & 6.379                     & 0.104                           & \multicolumn{1}{c|}{0.949}                                  & 13.0          \\
\multicolumn{1}{c|}{IID-SfMLearner\cite{iidsfm}}                                    & \multicolumn{1}{c|}{2024}                      & 0.058                          & 0.435                         & 4.820                     & 0.080                           & \multicolumn{1}{c|}{0.969}                                  & 14.8          \\
\multicolumn{1}{c|}{DVSMono \cite{dvs}}                                             & \multicolumn{1}{c|}{2024}                      & 0.055                          & 0.410                         & 4.797                     & 0.078                           & \multicolumn{1}{c|}{0.975}                                  & 27.0          \\
\multicolumn{1}{c|}{DARES \cite{dares}}                                             & \multicolumn{1}{c|}{2024}                      & 0.052                          & 0.356                         & 4.483                     & 0.073                           & \multicolumn{1}{c|}{0.980}                                  & 2.9           \\
\multicolumn{1}{c|}{EndoDAC\cite{endodac}}                                          & \multicolumn{1}{c|}{2024}                    & 0.052                          & 0.362                         & 4.464                     & 0.072                           & \multicolumn{1}{c|}{0.979}                                  & 1.6           \\
\multicolumn{1}{c|}{SfM-Diffusion \cite{sdiff}}                                      & \multicolumn{1}{c|}{2025}                    & 0.049                          & 0.366                         & 4.305                     & 0.078                           & \multicolumn{1}{c|}{0.975}                                  & 12.3          \\
\multicolumn{1}{c|}{MonoPCC\cite{pcc}}                                              & \multicolumn{1}{c|}{2025}                     & 0.051                          & 0.349                         & 4.488                     & 0.072                           & \multicolumn{1}{c|}{0.983}                                  & 27.0          \\ \hline
\multicolumn{1}{c|}{{EndoGMDE}}          & \multicolumn{1}{c|}{\multirow{2}{*}{Ours}} & \multirow{2}{*}{\begin{tabular}[c]{@{}c@{}}\textbf{0.047}\\ \color[HTML]{FE0000}{↓4.1\%}\end{tabular}}                 & \multirow{2}{*}{\begin{tabular}[c]{@{}c@{}}\textbf{0.307}\\ \color[HTML]{FE0000}{↓12\%}\end{tabular}}                & \multirow{2}{*}{\begin{tabular}[c]{@{}c@{}}\textbf{4.206}\\ \color[HTML]{FE0000}{↓2.3\%}\end{tabular}}            & \multirow{2}{*}{\begin{tabular}[c]{@{}c@{}}\textbf{0.067}\\ \color[HTML]{FE0000}{↓6.9\%}\end{tabular}}                  & \multicolumn{1}{c|}{\multirow{2}{*}{\begin{tabular}[c]{@{}c@{}}\textbf{0.985}\\ \color[HTML]{FE0000}{↑0.2\%}\end{tabular}}} & \multirow{2}{*}{3.7}           \\
\multicolumn{1}{c|}{$\Delta$} & \multicolumn{1}{c|}{} & & & & & \multicolumn{1}{c|}{} & \\ \hline
\end{tabular}
\label{comp1}
\end{table*}

\begin{table*}[]
\centering
\fontsize{8}{10}\selectfont
\renewcommand{\arraystretch}{1}
\caption{Comparison Results of Zero-shot Depth Estimation on Hamlyn and SERV-CT Dataset. \textbf{The BEST results} are in BOLD.}
\begin{tabular}{c|ccccc|ccccc}
\hline 
                                   & \multicolumn{5}{c|}{Hamlyn}                                                                                                                              & \multicolumn{5}{c}{SERV-CT}                                                                                                                              \\ \cline{2-11} 
\multirow{-2}{*}{\textbf{Methods}} & \textbf{$Rel_{Abs}$} & \textbf{$Rel_{Sq}$} & \textbf{$RMSE$} & \textbf{$RMSE_{Log}$} & \textbf{$\delta$} & \textbf{$Rel_{Abs}$} & \textbf{$Rel_{Sq}$} & \textbf{$RMSE$} & \textbf{$RMSE_{Log}$} & \textbf{$\delta$} \\ \hline
AF-SfMLearner\cite{afsfm}          & 0.168                          & 4.440                         & 13.870                    & 0.204                           & 0.770                     & 0.105                          & 1.772                         & 11.590                    & 0.134                           & 0.889                     \\
MonoViT \cite{mv}                  & 0.193                          & 10.512                        & 18.028                    & 0.220                           & 0.769                     & 0.103                          & 1.566                         & 11.482                    & 0.136                           & 0.895                     \\
LiteMono \cite{lm}                 & 0.179                          & 6.366                         & 15.196                    & 0.216                           & 0.754                     & 0.124                          & 2.314                         & 13.156                    & 0.175                           & 0.820                     \\
Depth Anything\cite{da}            & 0.154                          & 3.616                         & 12.733                    & 0.189                           & 0.784                     & 0.082                          & 1.122                         & 9.409                     & 0.110                           & 0.929                     \\
Depth Anything v2\cite{da2}        & 0.182                          & 4.994                         & 15.067                    & 0.219                           & 0.740                     & 0.092                          & 1.328                         & 10.407                    & 0.120                           & 0.918                     \\
IID-SfMLearner\cite{iidsfm}        & 0.171                          & 4.526                         & 14.066                    & 0.206                           & 0.767                     & 0.112                          & 1.956                         & 12.193                    & 0.138                           & 0.878                     \\
DVSMono\cite{dvs}                  & 0.143                          & \textbf{3.056}                         & 11.905                    & 0.181                           & 0.796                     & 0.095                          & 1.412                         & 10.811                    & 0.127                           & 0.907                     \\
DARES\cite{dares}                  & 0.164                          & 4.456                         & 13.625                    & 0.201                           & 0.783                     & 0.084                          & 1.179                         & 9.932                     & 0.113                           & 0.930                     \\
EndoDAC\cite{afsfm}                & 0.156                          & 3.854                         & 12.936                    & 0.193                           & 0.791                     & 0.084                          & 1.223                         & 9.683                     & 0.113                           & 0.922                     \\
MonoPCC\cite{pcc}                  & 0.158                          & 3.889                         & 13.205                    & 0.194                           & 0.782                     & 0.091                          & 1.265                         & 10.123                    & 0.117                           & 0.915                     \\ \hline
\multicolumn{1}{c|}{EndoGMDE}           & \multirow{2}{*}{\begin{tabular}[c]{@{}c@{}}\textbf{0.140}\\ \color[HTML]{FE0000}{↓2.1\%}\end{tabular}}                 & \multirow{2}{*}{\begin{tabular}[c]{@{}c@{}}3.092\\ \color[HTML]{32CB00}{↑1.2\%}\end{tabular}}                & \multirow{2}{*}{\begin{tabular}[c]{@{}c@{}}\textbf{11.749}\\ \color[HTML]{FE0000}{↓1.3\%}\end{tabular}}            & \multirow{2}{*}{\begin{tabular}[c]{@{}c@{}}\textbf{0.175}\\ \color[HTML]{FE0000}{↓3.3\%}\end{tabular}}                  & \multicolumn{1}{c|}{\multirow{2}{*}{\begin{tabular}[c]{@{}c@{}}\textbf{0.808}\\ \color[HTML]{FE0000}{↑1.5\%}\end{tabular}}} & \multirow{2}{*}{\begin{tabular}[c]{@{}c@{}}\textbf{0.069}\\ \color[HTML]{FE0000}{↓16\%}\end{tabular}}                 & \multirow{2}{*}{\begin{tabular}[c]{@{}c@{}}\textbf{0.307}\\ \color[HTML]{FE0000}{↓35\%}\end{tabular}}                & \multirow{2}{*}{\begin{tabular}[c]{@{}c@{}}\textbf{7.597}\\ \color[HTML]{FE0000}{↓19\%}\end{tabular}}            & \multirow{2}{*}{\begin{tabular}[c]{@{}c@{}}\textbf{0.088}\\ \color[HTML]{FE0000}{↓20\%}\end{tabular}}                  & \multirow{2}{*}{\begin{tabular}[c]{@{}c@{}}\textbf{0.963}\\ \color[HTML]{FE0000}{↑3.5\%}\end{tabular}}           \\
\multicolumn{1}{c|}{$\Delta$} & & & & & \multicolumn{1}{c|}{} & & & & & \\ \hline
\end{tabular}
\label{comp2}
\end{table*}

\subsection{Datasets}
\textbf{SCARED}: SCARED\cite{scared} dataset consists of 35 videos collected from porcine cadavers using a da Vinci Xi surgical robot. Following \cite{afsfm}, this dataset is divided into 15351, 1705, and 551 frames for training, validation, and test, respectively.

\textbf{Hamlyn}: Hamlyn\cite{edm} dataset contains numerous endoscopic videos of diverse robotic surgeries. In the experiments, all 92,672 endoscopic frames from the 21 rectified videos are used for zero-shot depth estimation.

\textbf{SERV-CT}: SERV-CT\cite{servct} dataset consists of 8 keyframes collected from two ex-vivo porcine cadavers. In the experiments, all 32 endoscopic frames are used for zero-shot depth estimation.

\textbf{SimCol}: SimCol\cite{simcol} dataset contains over 36,000 colonoscopic images based on the simulation in Unity3D. Following official website, the dataset is split into 28,776 and 9,009 frames for training and test, respectively.

\textbf{C3VD}: C3VD\cite{c3vd} dataset consists of 22 short video sequences collected from high-fidelity colon models by a clinical colonoscope. Following previous works \cite{c3vd-s}, 2889 frames from 8 videos are selected for zero-shot depth estimation.

\textbf{EndoMapper}: EndoMapper\cite{em} dataset consists of realistic colonoscopic sequences collected from regular medical practice. For sim-to-real evaluation, 2 frames are selected for the display of zero-shot depth estimation.

\subsection{Depth Estimation Results}

\begin{figure*}
    \centering
    \includegraphics[width=\linewidth]{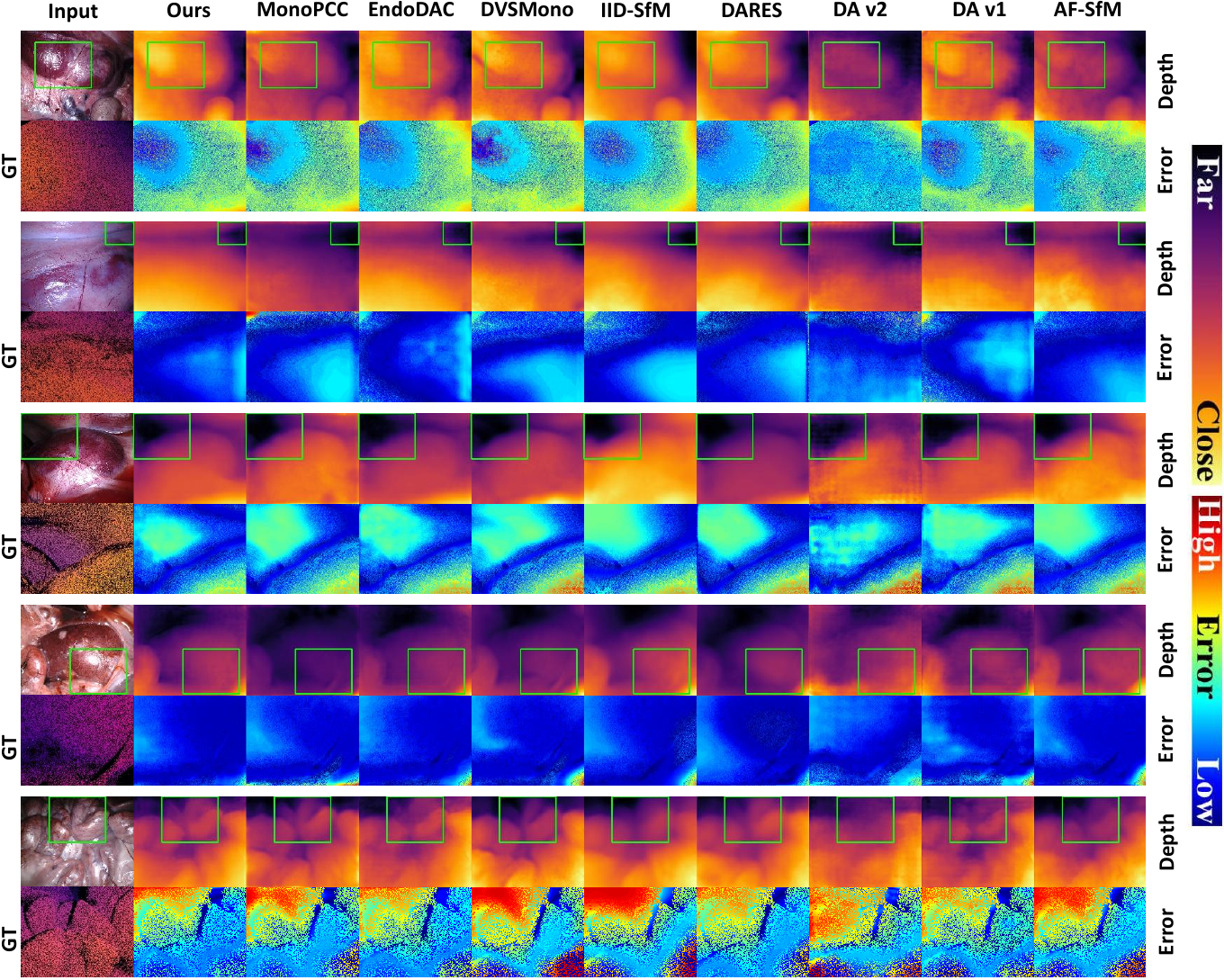}
    \caption{Qualitative results of depth estimation on SCARED dataset. The error maps denotes the relative error distribution in the area marked by the green box. The first image at the row of error maps is the ground truth of depth map in corresponding area.}
    \label{vis1}
\end{figure*}

\begin{table*}[!ht]
\centering
\fontsize{8}{10}\selectfont
\renewcommand{\arraystretch}{1}
\caption{Comparison Results of Depth Estimation on SimCol and C3VD Dataset. \textbf{The BEST results} are in BOLD.}
\begin{tabular}{c|ccccc|ccccc}
\hline
                                   & \multicolumn{5}{c|}{\textbf{SimCol (Trained)}}                                                           & \multicolumn{5}{c}{\textbf{C3VD (Zero-shot)}}                                                            \\ \cline{2-11} 
\multirow{-2}{*}{\textbf{Methods}} & \textbf{$Rel_{Abs}$} & \textbf{$Rel_{Sq}$} & \textbf{$RMSE$} & \textbf{$RMSE_{Log}$} & \textbf{$\delta$} & \textbf{$Rel_{Abs}$} & \textbf{$Rel_{Sq}$} & \textbf{$RMSE$} & \textbf{$RMSE_{Log}$} & \textbf{$\delta$} \\ \hline
AF-SfMLearner\cite{afsfm}          & 0.086                & 0.080               & 0.426           & 0.109                 & 0.950             & 0.142                & 1.495               & 7.897           & 0.188                 & 0.804             \\
IID-SfMLearner\cite{iidsfm}        & 0.080                & 0.069               & 0.429           & \textbf{0.108}        & \textbf{0.959}    & 0.140                & 1.484               & 7.733           & 0.186                 & 0.810             \\
EndoDAC\cite{endodac}              & 0.107                & 0.110               & 0.530           & 0.133                 & 0.924             & 0.103                & 0.652               & 5.306           & 0.127                 & 0.918             \\ \hline
\multicolumn{1}{c|}{EndoGMDE}           & \multirow{2}{*}{\begin{tabular}[c]{@{}c@{}}\textbf{0.078}\\ \color[HTML]{FE0000}{↓2.5\%}\end{tabular}}                 & \multirow{2}{*}{\begin{tabular}[c]{@{}c@{}}0.056\\ \color[HTML]{FE0000}{↓19\%}\end{tabular}}                & \multirow{2}{*}{\begin{tabular}[c]{@{}c@{}}\textbf{0.421}\\ \color[HTML]{FE0000}{↓1.2\%}\end{tabular}}            & \multirow{2}{*}{\begin{tabular}[c]{@{}c@{}}\textbf{0.109}\\ \color[HTML]{32CB00}{↑0.9\%}\end{tabular}}                  & \multicolumn{1}{c|}{\multirow{2}{*}{\begin{tabular}[c]{@{}c@{}}\textbf{0.959}\\ 0\%\end{tabular}}} & \multirow{2}{*}{\begin{tabular}[c]{@{}c@{}}\textbf{0.096}\\ \color[HTML]{FE0000}{↓6.8\%}\end{tabular}}                 & \multirow{2}{*}{\begin{tabular}[c]{@{}c@{}}\textbf{0.635}\\ \color[HTML]{FE0000}{↓2.6\%}\end{tabular}}                & \multirow{2}{*}{\begin{tabular}[c]{@{}c@{}}\textbf{5.241}\\ \color[HTML]{FE0000}{↓1.2\%}\end{tabular}}            & \multirow{2}{*}{\begin{tabular}[c]{@{}c@{}}\textbf{0.120}\\ \color[HTML]{FE0000}{↓5.5\%}\end{tabular}}                  & \multirow{2}{*}{\begin{tabular}[c]{@{}c@{}}\textbf{0.923}\\ \color[HTML]{FE0000}{↑0.5\%}\end{tabular}}           \\
\multicolumn{1}{c|}{$\Delta$} & & & & & \multicolumn{1}{c|}{} & & & & & \\ \hline
\end{tabular}
\label{comp3}
\end{table*}

\subsubsection{Results on Realistic Data}
\begin{figure*}
    \centering
    \includegraphics[width=0.95\linewidth]{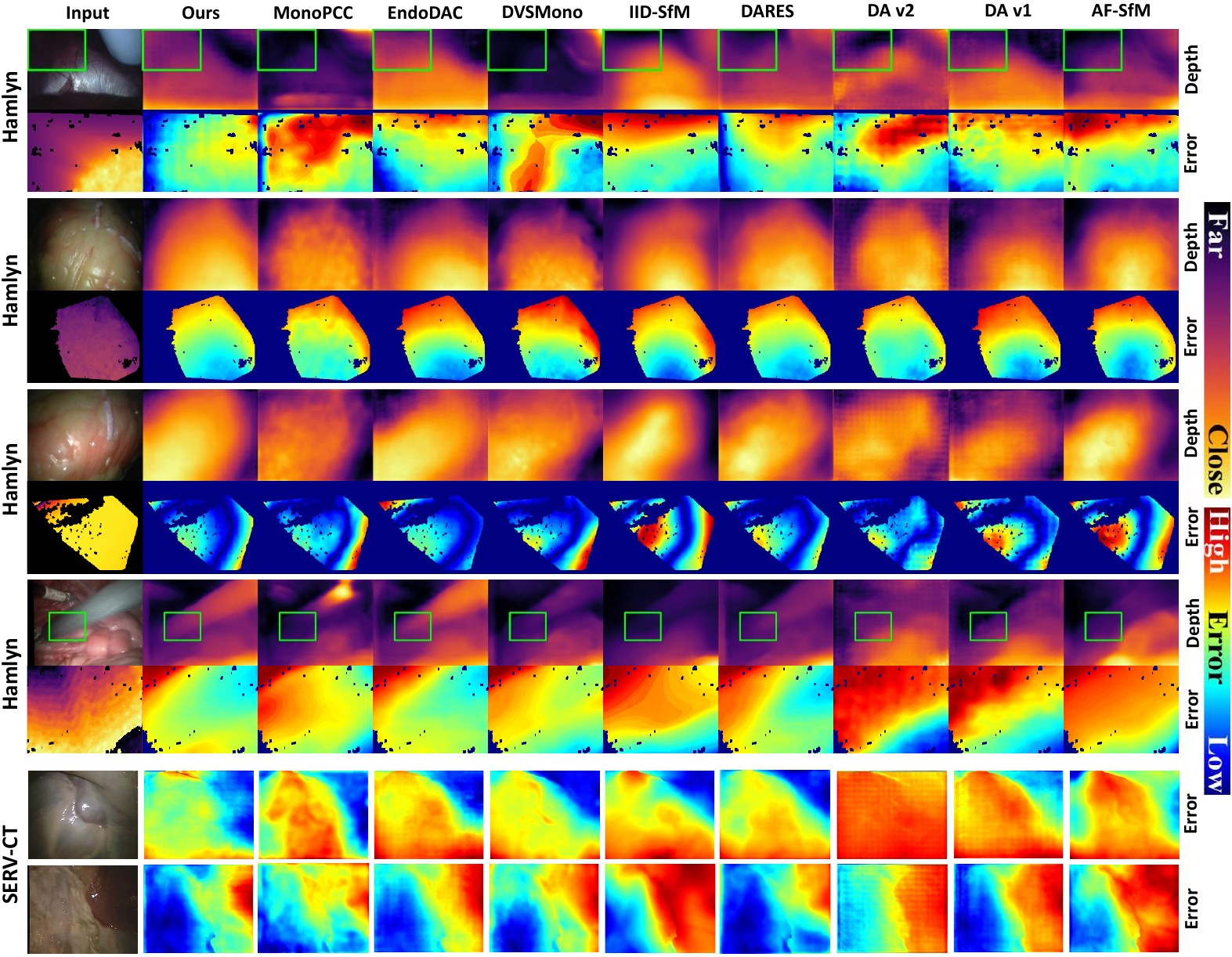}
    \caption{Qualitative results of zero-shot depth estimation on Hamlyn and SERV-CT dataset. The error maps denotes the relative error distribution in the area marked by the green box. Note that if there is no green box, the error map denotes the errors in the whole picture. The image below the input raw image is the ground truth of depth map in corresponding area.}
    \label{vis2}
\end{figure*}

The proposed method is evaluated on three different realistic datasets, SCARED, Hamlyn and SERV-CT. Meanwhile, our method is compared with state-of-the-art works including methods for endoscopic scenes (MonoPCC\cite{pcc}, EndoDAC\cite{endodac}, DVSMono\cite{dvs}, DARES\cite{dares}, SfM-Diffusion\cite{sdiff}, IID-SfM\cite{iidsfm}, Surgical-DINO\cite{sdino}, AF-SfM\cite{afsfm} and method from \cite{tmi}) and natural scenes (MonoViT\cite{mv}, LiteMono\cite{lm} and Depth Anything\cite{da,da2}). All frameworks are trained and tested on SCARED dataset at first, results of which are shown in Table. \ref{comp1} and Fig. \ref{vis1}. To evaluate the generalizability of the model, all frameworks trained on SCARED dataset are utilized to perform zero-shot depth estimation on Hamlyn and SERV-CT dataset, and the experimental results are shown in Table. \ref{comp2} and Fig. \ref{vis2}. Note that some works without published codes and weights, including Surgical-DINO\cite{sdino}, Yang's method\cite{tmi} and SfM-Diffusion\cite{sdiff}, are not compared for zero-shot estimation.

According the quantitative results on multiple datasets, the proposed method outperforms existing methods with lower error ($Rel$ and $RMSE$) and higher accuracy ($\delta$) in diverse endoscopic scenes. The best generalization of our method is also demonstrated based on the zero-shot evaluation. Based on qualitative results, the proposed method can provide more accurate depth estimation for the edge of tissues or instruments and smoother depth estimation for the tissue surface without interference of the light.

\begin{figure*}[!ht]
    \centering
    \includegraphics[width=0.95\linewidth]{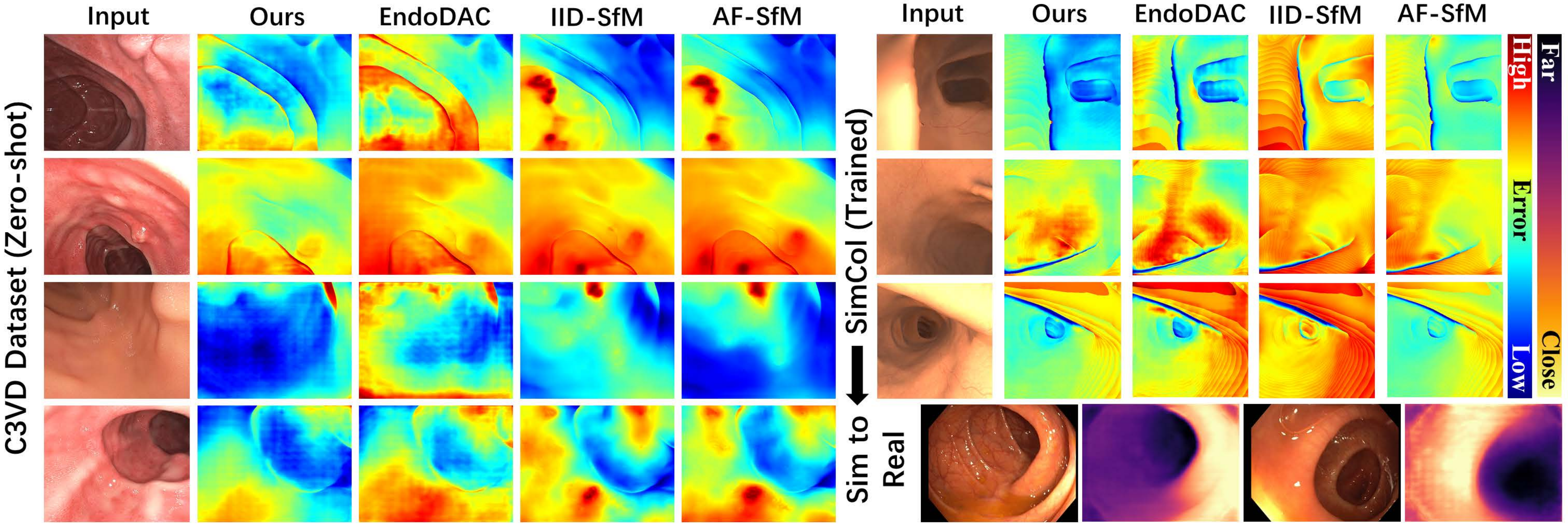}
    \caption{Qualitative results of depth estimation on Simulated dataset and sim-to-real test.}
    \label{vis3}
\end{figure*}

\subsubsection{Results on Simulated Data and Sim-to-Real}
To further demonstrate the performance of our method, the proposed framework is trained and evaluated on simulated datasets. Firstly, our method is compared with existing methods, which are with public training codes, on SimCol dataset. Meanwhile, the models trained on SimCol dataset are compared with previous works for zero-shot depth estimation on C3VD dataset. Furthermore, the proposed model with the pretrained weights from SimCol is also utilized to perform depth estimation on samples from EndoMapper, which can display the sim-to-real performance of our work. The quantitative results are shown in Table \ref{comp3}, while qualitative results including sim-to-real depth estimation are shown in Fig. \ref{vis3}. According to the experimental results, the outstanding performance and generalization of the proposed method are further proved.

\begin{table*}[!ht]
\centering
\fontsize{8}{8}\selectfont
\renewcommand{\arraystretch}{1.2}
\caption{Ablations on Parameter-efficient Finetuning Strategy. {\color[HTML]{FE0000}The best results} are in red.}
\begin{tabular}{cc|ccccc}
\hline
\multicolumn{2}{c|}{\textbf{PEFT}}          & \multicolumn{5}{c}{\textbf{SCARED  /    SERV-CT  /  Hamlyn}}                                                                                                         \\ \hline
\textbf{QKV}          & \textbf{FFN}        & \textbf{$Rel_{Abs}\downarrow$} & \textbf{$Rel_{Sq}\downarrow$}  & \textbf{$RMSE\downarrow$}       & \textbf{$RMSE_{Log}\downarrow$} & \textbf{$\delta\uparrow$}      \\ \hline
\rowcolor[HTML]{EFEFEF} 
\multicolumn{2}{c|}{\cellcolor[HTML]{EFEFEF}BW-MoLE} & {\color[HTML]{FE0000} 0.047 / 0.069 / 0.140} & {\color[HTML]{FE0000} 0.307 / 0.730} / 3.092 & {\color[HTML]{FE0000} 4.206 / 7.597 / 11.749} & {\color[HTML]{FE0000} 0.067 / 0.088 / 0.176} & {\color[HTML]{FE0000} 0.985 / 0.963 / 0.808} \\
\multicolumn{2}{c|}{GW-MoLE\cite{mola}}              & 0.049 / 0.083 / 0.144          & 0.326 / 1.039 / 3.391          & 4.303 / 9.163 / 12.034          & 0.069 / 0.107 / 0.179           & 0.983 / 0.935 / 0.808          \\
MoLE(2/3)             & MoLE(2/5)           & 0.048 / 0.083 / 0.145          & 0.315 / 1.081 / 3.345          & 4.210 / 8.948 / 12.052           & 0.068 / 0.105 / 0.180           & 0.984 / 0.936 / 0.807          \\
MoLE(1/3)             & MoLE(2/5)           & 0.050 / 0.075 / 0.147           & 0.364 / 0.854 / 3.416          & 4.556 / 8.240 / 12.257          & 0.072 / 0.096 / 0.182           & 0.981 / 0.950 / 0.800          \\
MoLE(2/4)             & MoLE(2/5)           & 0.051 / 0.083 / 0.153          & 0.348 / 1.107 / 3.672          & 4.428 / 9.346 / 12.721          & 0.072 / 0.108 / 0.189           & 0.980 / 0.937 / 0.792           \\
MoLE(2/4)             & MoLE(2/4)           & 0.052 / 0.083 / 0.156          & 0.392 / 1.068 / 3.973          & 4.666 / 8.989 / 12.941          & 0.074 / 0.106 / 0.192           & 0.980 / 0.937 / 0.792           \\ \hline
BW-MoLE             & $\times$        & 0.051 / 0.084 / 0.147          & 0.364 / 1.061 / 3.363          & 4.511 / 9.223 / 12.212          & 0.073 / 0.107 / 0.182           & 0.979 / 0.941 / 0.799          \\
$\times$          & BW-MoLE           & 0.050 / 0.082 / 0.153           & 0.364 / 1.066 / 3.694          & 4.528 / 9.299 / 12.843          & 0.071 / 0.108 / 0.189           & 0.981 / 0.931 / 0.790          \\
$\times$          & MoLE(2/5)           & 0.055 / 0.079 / 0.156          & 0.476 / 1.065 / 4.079          & 5.119 / 9.094 / 13.062          & 0.081 / 0.105 / 0.192           & 0.977 / 0.926 / 0.793          \\ \hline
\multicolumn{2}{c|}{HydraLoRA\cite{hdlora}} & 0.052 / 0.098 / 0.155          & 0.361 / 1.675 / 4.066          & 4.525 / 10.931 / 12.986         & 0.073 / 0.125 / 0.191           & 0.982 / 0.904 / 0.796          \\
\multicolumn{2}{c|}{AdaLoRA\cite{adlora}}   & 0.054 / 0.076 / 0.144          & 0.438 / 0.985 / {\color[HTML]{FE0000}3.057}          & 4.949 / 8.992 / 12.075          & 0.078 / 0.103 / 0.179           & 0.978 / 0.933 / 0.801          \\
\multicolumn{2}{c|}{DV-LoRA\cite{endodac}}  & 0.050 / 0.086 / 0.151           & 0.334 / 1.172 / 3.639          & 4.342 / 9.403 / 12.469          & 0.071 / 0.108 / 0.186           & 0.983 / 0.929 / 0.798          \\
\multicolumn{2}{c|}{LoRA\cite{lora}}        & 0.049 / 0.081 / 0.150          & 0.336 / 1.065 / 3.681          & 4.381 / 9.035 / 12.508          & 0.070 / 0.103 / 0.185            & 0.983 / 0.945 / 0.800          \\ \hline
\end{tabular}
\label{ab1}
\end{table*}

\begin{table*}[!ht]
\centering
\fontsize{8}{8}\selectfont
\renewcommand{\arraystretch}{1.2}
\caption{Ablation studies on Self-supervised Learning (SSL) pipeline. {\color[HTML]{FE0000}The best results} are in red.}
\begin{tabular}{c|ccccc}
\hline
\multirow{2}{*}{\textbf{SSL   Pipeline}} & \multicolumn{5}{c}{\textbf{SCARED / SERV-CT  /    Hamlyn}}                                                                                                           \\ \cline{2-6} 
                                         & \textbf{$Rel_{Abs}\downarrow$} & \textbf{$Rel_{Sq}\downarrow$}  & \textbf{$RMSE\downarrow$}       & \textbf{$RMSE_{Log}\downarrow$} & \textbf{$\delta\uparrow$}      \\ \hline
\rowcolor[HTML]{EFEFEF} 
Ours                                      & {\color[HTML]{FE0000} 0.047 / 0.069 / 0.140} & {\color[HTML]{FE0000} 0.307 / 0.730 / 3.092} & {\color[HTML]{FE0000} 4.206 / 7.597 / 11.749} & {\color[HTML]{FE0000} 0.067 / 0.088} / 0.176 & {\color[HTML]{FE0000} 0.985} / 0.963 / 0.808 \\
w/o $\mathcal{L}_{if}$                   & 0.048 / 0.070 / 0.141          & 0.332 / 0.766 / 3.109          & 4.350 / 7.723 / 11.774           & 0.069 / 0.089 / {\color[HTML]{FE0000}0.175 }          & 0.983 / {\color[HTML]{FE0000}0.966 / 0.810}          \\
w/o $\mathcal{L}_{ret}$                 & 0.050 / 0.082 / 0.149           & 0.346 / 1.004 / 3.508          & 4.455 / 8.924 / 12.470          & 0.071 / 0.104 / 0.184           & 0.983 / 0.942 / 0.798          \\
$\mathcal{L}_{rec}$  w/o $A_{n\times}$                   & 0.050 / 0.076 / 0.142           & 0.335 / 0.904 / 3.138          & 4.344 / 8.562 / 11.842          & 0.070 / 0.097/ 0.180             & 0.980 / 0.959 / 0.805           \\ \hline
AF-SfMLearner\cite{afsfm}                & 0.049 / 0.083 / 0.142          & 0.339 / 1.071 / 3.138          & 4.390 / 9.086 / 11.842           & 0.070 / 0.107 / 0.180            & 0.983 / 0.932 / 0.805          \\
IID-SfMLearner\cite{iidsfm}              & 0.058 / 0.092 / 0.148          & 0.448 / 1.342 / 3.302          & 4.997 / 9.924 / 12.372          & 0.081 / 0.118 / 0.184           & 0.976 / 0.913 / 0.794          \\ \hline
\end{tabular}
\label{ab2}
\end{table*}

\subsection{Ablation Studies}
\subsubsection{Ablations on Parameter-efficient Finetuning} Ablation studies are performed to demonstrate the effect of the proposed parameter-efficient finetuning based on block-wise mixture of low-rank experts (BW-MoLE), as Table \ref{ab1} shows. Firstly, the allocation of the experts are replaced with group-wise allocation (GW-MoLE\cite{mola}) and average allocation. Note that the deeper group is also allocated with more experts in GW-MoLE according to Fig. \ref{allo}. The numbers of experts are set to 2, 4, 6 and 8 for each group consisting of four Transformer blocks for group-wise allocation, while the top-ks in all blocks are equal to 2. Moreover, the number of experts is set to x with the top-k setting, represented as \textit{MoLE(k/x)}, for average allocation. Furthermore, the finetuning of each linear layer is demonstrated based on ablation studies. At last, the finetuning based on the proposed method is also compared with the PEFT with multiple kinds of LoRA.
\subsubsection{Ablations on Self-supervised Learning} Ablation studies are performed to demonstrate the effect of the proposed self-supervised training strategy, as Table \ref{ab2} shows. For our training strategy, the experiments are implemented to prove the effect of the loss from augmented image, Retinex algorithm and illumination-free alignment. Moreover, the proposed model is also trained based on the existing training pipeline AF-SfMLearner \cite{afsfm} and IID-SfMLearner \cite{iid} for comparison.

\begin{figure}
    \centering
    \includegraphics[width=\linewidth]{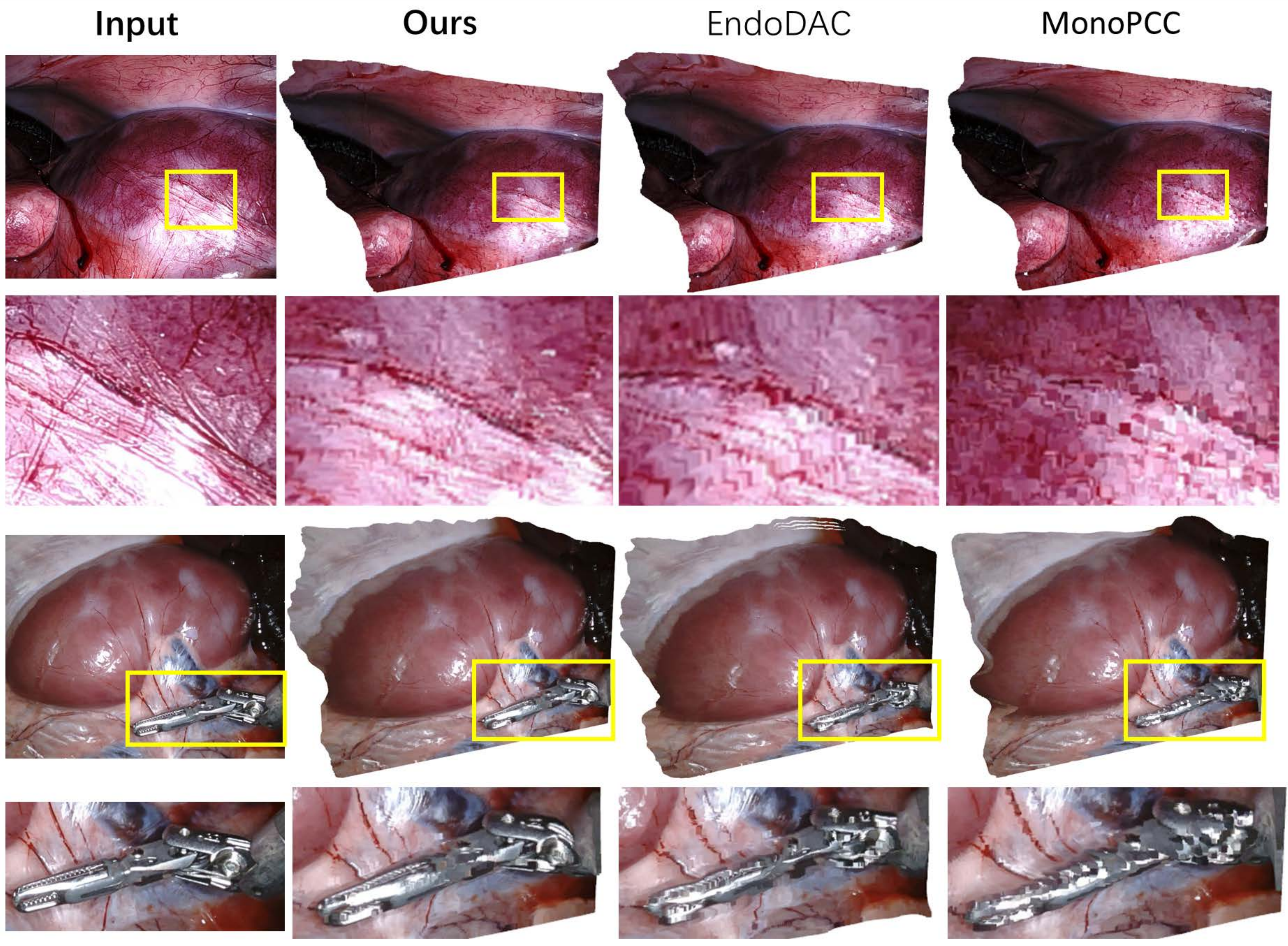}
    \caption{3D scene reconstruction results on data from SCARED dataset.}
    \label{rec}
\end{figure}

\subsection{3D Reconstruction Results}
3D reconstruction is a significant application of depth estimation for endoscopy. Therefore, we also perform 3D scene reconstruction based on the estimated depth maps using Open3D (https://www.open3d.org/). As Fig. \ref{rec} shows, the reconstructed scenes based on the depth maps provided by our method is with more clear and accuracy texture and appearance, compared with results from \cite{endodac} and \cite{pcc}.

\begin{figure*}
    \centering
    \includegraphics[width=\linewidth]{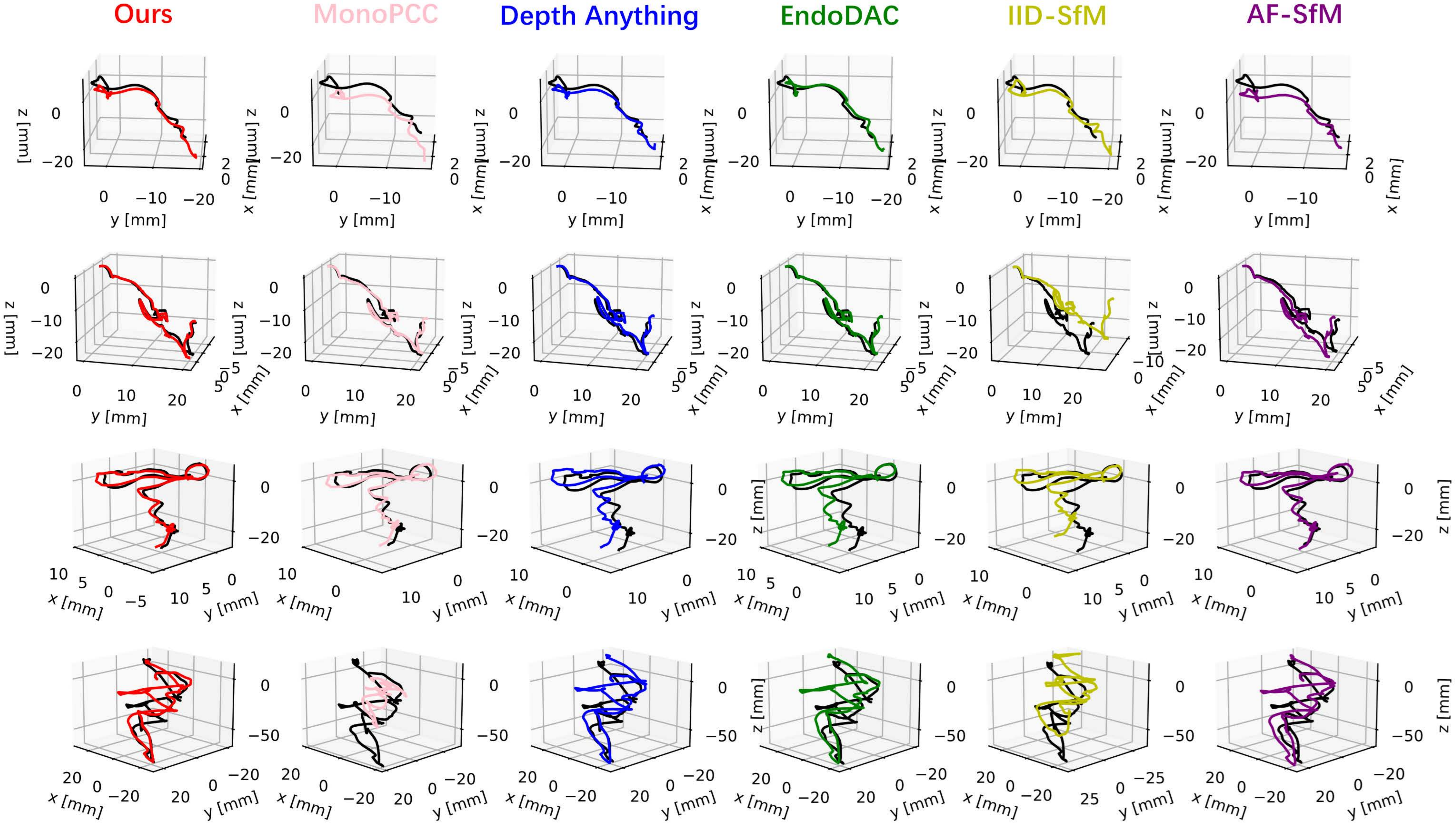}
    \caption{Qualitative results of ego-motion estimation on SCARED dataset. Black lines denote the ground truths.}
    \label{ego}
\end{figure*}

\subsection{Ego-motion Estimation Results}
Ego-motion estimation, as one of important part in the self-supervised framework, is also evaluated with experiments. Four sequences from the test split of SCARED dataset are selected to perform ego-motion estimation. The accuracy of ego-motion estimation is evaluated and compared based on Absolute Trajectory Error (ATE), following previous works. Based on experimental results in Table \ref{comp4} and Fig. \ref{ego}, the proposed method can perform the most accurate ego-motion estimation, compared with multiple existing methods including MonoPCC\cite{pcc}, EndoDAC\cite{endodac}, IID-SfM\cite{iidsfm} and AF-SfM\cite{afsfm}.

\begin{table}[]
\centering
\fontsize{8}{10}\selectfont
\renewcommand{\arraystretch}{1.2}
\caption{Ego-motion estimation results on SCARED dataset}
\begin{tabular}{c|cccc}
\hline
                                   & \multicolumn{4}{c}{\textbf{ATE$\downarrow$}}                                      \\ \cline{2-5} 
\multirow{-2}{*}{\textbf{Methods}} & \textbf{Seq. 1} & \textbf{Seq. 2} & \textbf{Seq. 3} & \textbf{Seq. 4} \\ \hline
AF-SfMLearner\cite{afsfm}          & 0.0941          & 0.0841          & \textbf{0.0742} & \underline{0.0682}    \\
Depth Anything\cite{da}            & \underline{0.0932}    & \underline{0.0815}    & 0.0794          & 0.0696          \\
IID-SfMLearner\cite{iidsfm}        & 0.0951          & 0.0851          & 0.0764          & \underline{0.0682}    \\
EndoDAC\cite{endodac}              & 0.0936          & 0.0832          & 0.0776          & 0.0687          \\
MonoPCC\cite{pcc}                  & 0.1040          & 0.0842          & 0.0781          & 0.0939          \\
\rowcolor[HTML]{EFEFEF} 
Ours                               & \textbf{0.0927} & \textbf{0.0814} & \underline{0.0754}    & \textbf{0.0678} \\ \hline
\end{tabular}
\label{comp4}
\end{table}

\section{Discussion}

\subsection{Outstanding Accuracy and Generalization}
The proposed method is compared with state-of-the-art works across different kinds of datasets, including data from different tissue during different procedures collected by different methods. The experimental results demonstrate the outstanding performance and generalization of the proposed method. Moreover, results of 3D reconstruction and ego-motion estimation further display the outstanding performance of the proposed method, compared with existing methods.

\begin{figure}
    \centering
    \includegraphics[width=\linewidth]{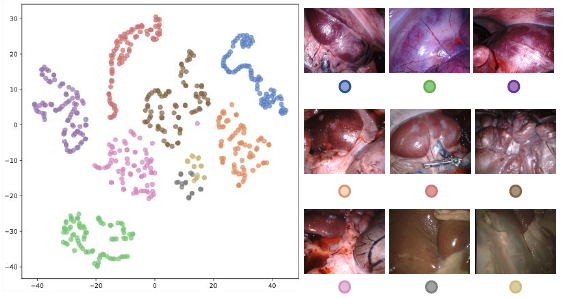}
    \caption{Visualization of the experts' weights based on t-SNE.}
    \label{dis}
\end{figure}

\subsection{Generalization on Diverse Scenes}

Based on foundation models, different from previous finetuning based on LoRA, mixture of experts is firstly introduced into the depth estimation model to boost the generalization by adapting weighted inference based on diverse input features. The effect of the block-wise allocated low-rank experts is demonstrated based on a series of ablation studies, compared with various LoRA and original MoE as well. Furthermore, based on 551 test subjects from 7 kinds of scenes in SCARED dataset and 32 subjects from 2 sequences in SERV-CT dataset, the predicted weights of experts are visualized using t-SNE in Fig. \ref{dis}. The visualization result demonstrates that the experts are adaptively selected based on the input features from different scenes.

\begin{figure}
    \centering
    \includegraphics[width=\linewidth]{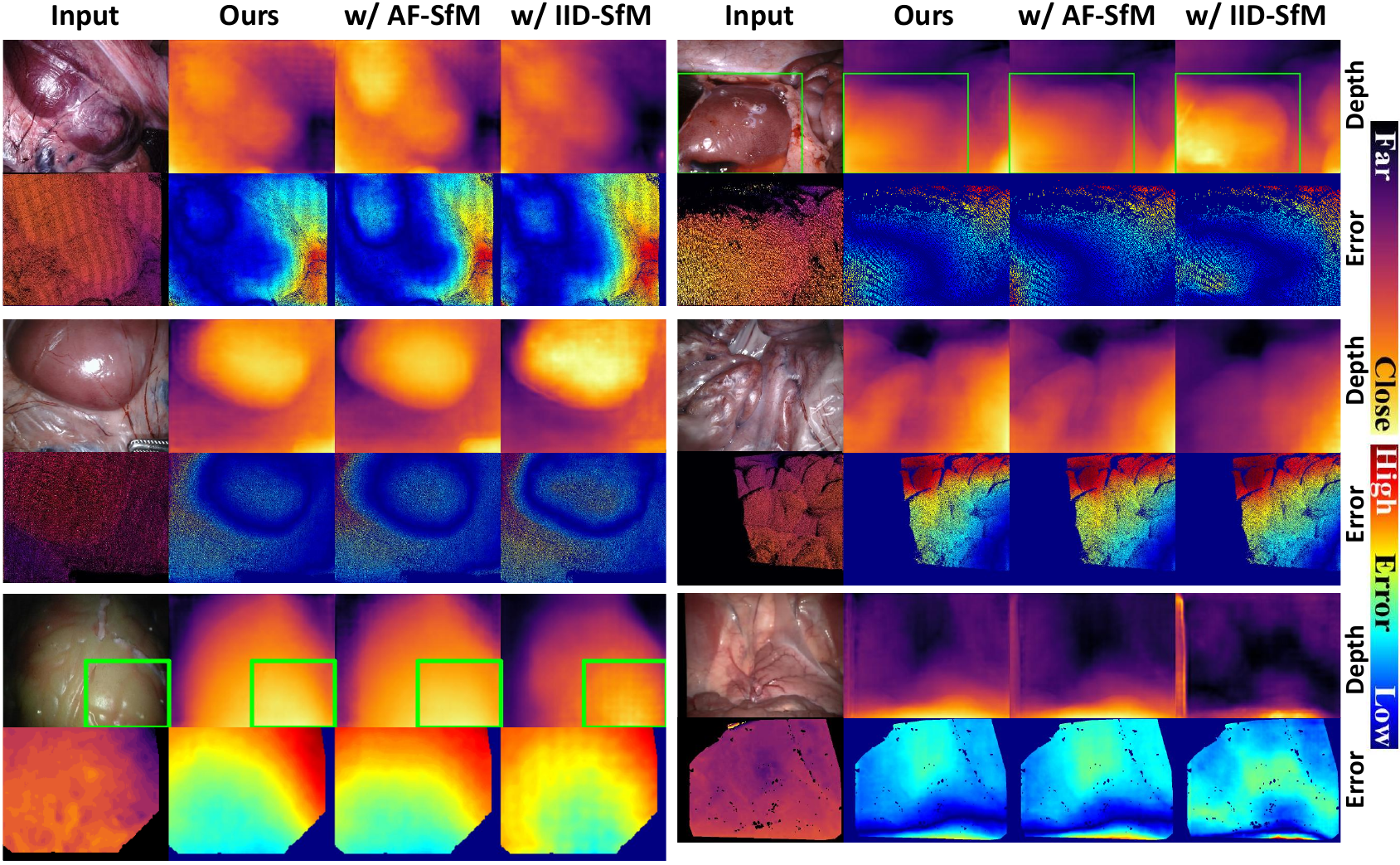}
    \caption{Comparison of results from models trained via different pipelines. The layout of this figure follows Fig.\ref{vis2}.}
    \label{vis_ab}
\end{figure}

\subsection{Generalization on Diverse Illumination}

Different from previous self-supervised training, both of the interference of reflectance and the brightness inconsistency are jointly dealt with by the proposed training framework. The superiority of our pipeline is proved based on qualitative results from ablation studies and comparison with existing pipelines. Furthermore, the proposed self-supervised training strategy is compared with previous pipelines on the sample from SCARED (first two rows) and Hamlyn (the last row) dataset in Fig. \ref{vis_ab}, which demonstrates the filtering of influence from light inconsistency by our method.

\begin{table}[]
\centering
\caption{Comparison of Inference Speed}
\begin{tabular}{c|cc}
\hline
\multirow{2}{*}{\textbf{Methods}}             & \multicolumn{2}{c}{\textbf{Speed(ms/frame)}} \\ \cline{2-3} 
                                              & \textbf{SCARED}       & \textbf{Hamlyn}      \\ \hline
Depth Anything\cite{da,da2}      & 5.0                   & 3.0                   \\
IID-SfMLearner\cite{iidsfm} & 2.0                   & 1.0                  \\
DVSMono\cite{dvs}            & 12.7                  & 10.5                 \\
EndoDAC\cite{endodac}        & 5.7                   & 4.0                  \\
MonoPCC\cite{pcc}            & 12.6                  & 10.5                 \\
\rowcolor[HTML]{EFEFEF}
Ours                                          & 34.9($\sim$30fps)                  & 32.5($>$30fps)                 \\ \hline
\end{tabular}
\label{speed}
\end{table}

\subsection{Limitations \& Future Works}
Although the proposed method achieves state-of-the-art performance compared with existing methods on monocular depth estimation in diverse endoscopic scenes, The inference speed of our method is lower than previous works while the speed still achieves about 30fps which could meet the real-time requirement, shown in Table.\ref{speed}. Although the number of training parameters remains low in the proposed framework, the step-by-step training strategy and the usage of foundation model still need lots of computational resources. In the future, more efficient training pipeline and foundation model could be developed for endoscopic depth estimation. Meanwhile, multimodal visual features such as semantic feature could boost the performance of depth estimation in endoscopy.

\section{Conclusion}
In this work, a novel self-supervised framework based on a novel parameter-efficient finetuning with mixture of experts is proposed for generalizable depth estimation in diverse endoscopic scenes. The experiments on multiple datasets demonstrate the state-of-the-art performance and generalization of the proposed pipeline in diverse endoscopic scenes. This work can contribute to the engineering application of 3D perception in endoscopy for minimally invasive surgery and diagnosis. Although the proposed method achieves satisfactory performance, the lower inference speed (about 20-30 fps) and the higher computational cost (about 23GB GPU memory for training) are still limitations which need future development.

\bibliographystyle{IEEEtran}
\bibliography{references}

\end{document}